%% file: clemente22icra.tex
\documentclass[letterpaper, 10pt, conference]{ieeeconf}

\IEEEoverridecommandlockouts
\usepackage[skins,theorems]{tcolorbox}
\usepackage{pgfplots}
\usepackage{adjustbox}
\usepackage{float}
\usepackage{hyperref}
\usepackage{tikz, tikz-3dplot}
\usetikzlibrary{calc, positioning, matrix}
\usepackage{graphicx}
\usepackage{amsmath,amsfonts,amssymb}
\usepackage{upgreek}
\usepackage{bm}
\usepackage[utf8]{inputenc}
\usepackage[none]{hyphenat}
\usepackage{subfig}
\usepackage{svg}
\usepackage[font=footnotesize,labelfont=bf,belowskip=-13pt]{caption}
\usepackage{acronym}
\usepackage[acronym,symbols,nogroupskip,nonumberlist,nomain,section=section]{glossaries-extra}
\usepackage{pdfpages}

\tcbset{highlight math style={enhanced,
  colframe=red,colback=white,arc=0pt,boxrule=1.5pt}}

\definecolor{cadmiumgreen}{rgb}{0.0, 0.42, 0.24}
\definecolor{deeppurple}{HTML}{673AB7}
\definecolor{amber}{HTML}{FFC107}

\setabbreviationstyle[acronym]{long-short}
\include{acronyms}

\tdplotsetmaincoords{70}{40}
\begin{document}
\title{
	Foothold Evaluation Criterion for Dynamic Transition Feasibility for 
	Quadruped Robots}

\author{Luca Clemente$^{1,2}$\thanks{$^{1}$ Dynamic Legged 
Systems (DLS) lab, Istituto 
			Italiano di Tecnologia (IIT), Genova (Italy). \textit{email} 
			firstname.lastname@iit.it},
Octavio Villarreal$^1$\thanks{$^2$ Department of Electronics and 
Telecommunications, Collegio di
	Ingegneria Informatica, del Cinema e Meccatronica, Politecnico di Torino,
	Turin (Italy). \textit{s265222@studenti.polito.it}},
Angelo Bratta$^1$,
Michele Focchi$^1$,
Victor Barasuol$^1$,\\
Giovanni Gerardo Muscolo$^3$\thanks{$^3$Department of Computer Science, 
University 
of Verona, Verona (Italy). \textit{giovannigerardo.muscolo@univr.it }}, 
and Claudio Semini$^1$}
\null
\includepdf[pages=-]{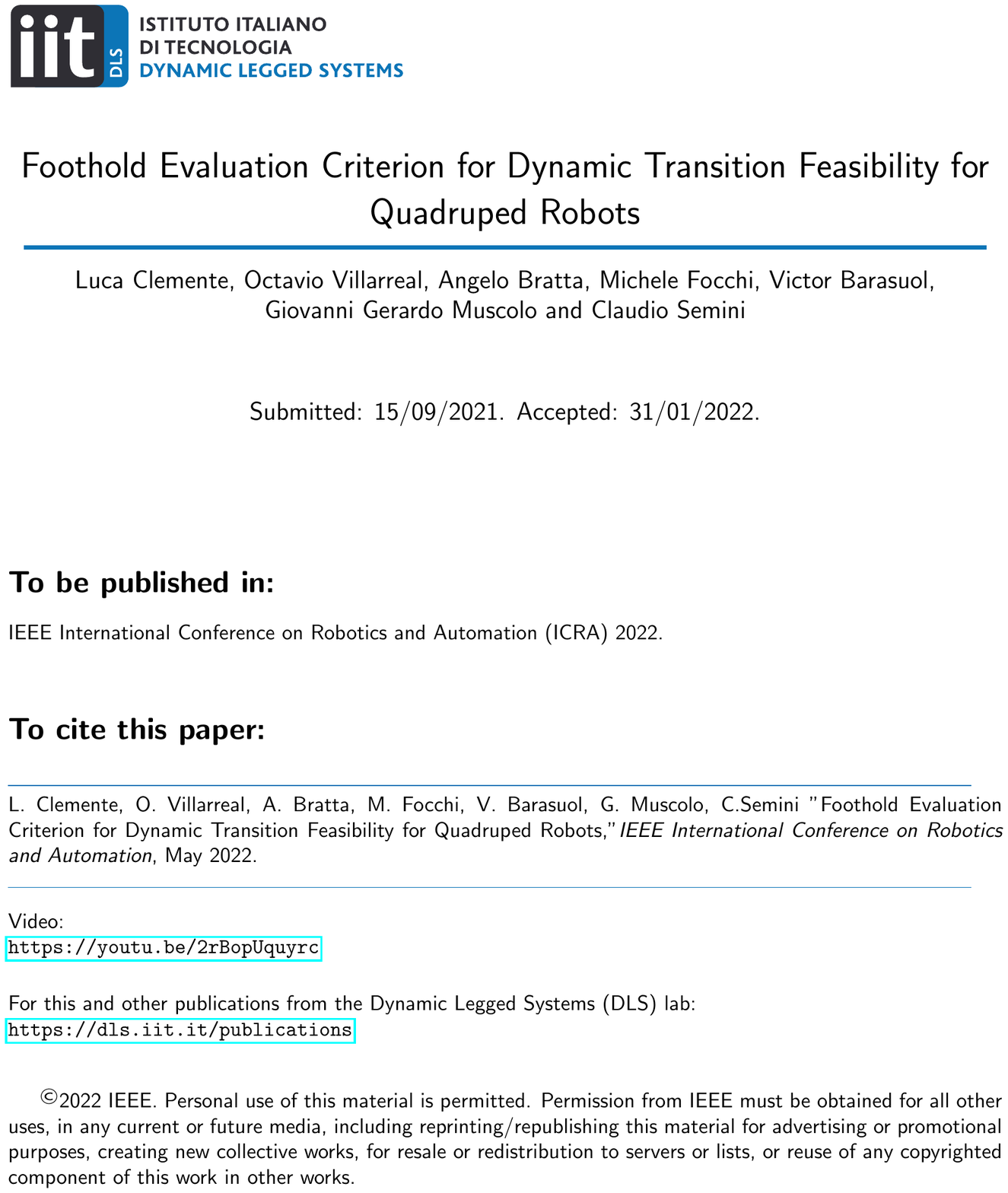}
\pgfdeclarelayer{bg}    
\pgfsetlayers{bg,main}

\maketitle
\begin{abstract}
To traverse complex scenarios reliably a legged robot needs to move its base 
aided by
the ground reaction forces, which can only be generated by the legs that are
momentarily in contact with the ground. A proper
selection of footholds is crucial for maintaining balance. In
this paper, we propose a foothold evaluation criterion that considers the
transition feasibility for both linear and angular dynamics to overcome complex
scenarios. We devise convex and nonlinear formulations as a direct extension
of \cite{CCROC} in a receding-horizon fashion to grant dynamic feasibility for
future behaviours. The criterion is integrated with a Vision-based Foothold
Adaptation (VFA) strategy that takes into account the robot kinematics, leg
collisions and terrain morphology. We verify the validity of the selected
footholds and the generated trajectories in simulation and experiments with the 90kg quadruped robot HyQ.
\end{abstract}
\IEEEpeerreviewmaketitle
\section{Introduction}
Legged robots are versatile machines that make use of their sensors to react, 
adapt and navigate in complex scenarios. 
To  move, the robot needs 
to choose a trajectory for 
the motion of its base, and 
decide a feasible 
contact sequence for its feet to follow the base. Motions and contacts need to 
be both dynamically (e.g., inside the torque limits of the robot's actuators or not falling)  
and  kinematically  feasible (e.g.,  inside  the  joints'  range  of  motion). 

Optimization-based techniques for legged robot whole-body control
\cite{kuindersma16auro, kim18iros,  DiCarlo2018, fahmi19ral} have become prevalent in locomotion, allowing robots to deal with complex terrain. 
Furthermore, some approaches explicitly include terrain 
information into the formulation of the locomotion problem 
\cite{aceituno18ral,Winkler2018,Villarreal2}. This is possible thanks to advances in 
state estimation and mapping 
\cite{nobilicamurri2017rss,flayols17humanoids,Fankhauser2016GridMapLibrary}, 
as well as the algorithmic developments such as the use of automatic 
differentiation \cite{giftthaler17ar} and differential dynamic programming 
\cite{mayne66ijc,li20ral}. However, in many cases the introduction of the 
terrain is not trivial, especially if kinematic and dynamic feasibility are 
considered. We categorize two main ways to tackle the problem: (a) 
coupled approaches, in which the base trajectory and footholds are optimized 
jointly, and (b) decoupled approaches, in which the footholds are selected 
first and then the trajectory is optimized to follow the footholds. 
\begin{figure}[!t]
	\centering
	\includegraphics[width=0.9\columnwidth]{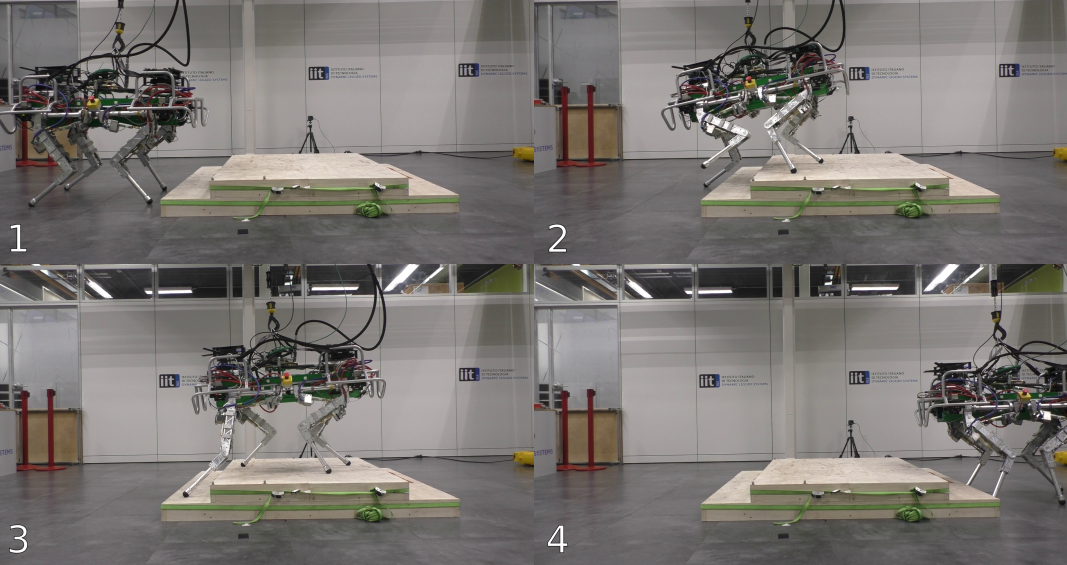}
	\caption{Snapshots of the HyQ robot climbing stair scenario using the 
	proposed foothold evaluation criterion for dynamic transition feasibility.}
	\label{fig:hyq_pallet_exp}
\end{figure}

The main advantage of coupled approaches is that the optimized footholds and 
trajectories are guaranteed to be realizable by the robot since kinematic and 
dynamic constraints can be enforced. The main disadvantage is that 
solving this problem involves a large number of variables and the 
nature of the problem is in general nonlinear with a large 
combinatorial space. 

Decoupled approaches split the problem in two. The foothold selection can be 
done heuristically or by solving a less computationally expensive optimization 
problem. This releases computational load from the main optimization. The 
disadvantage is that it is difficult to guarantee that the selected footholds 
yield dynamically feasible trajectory for the robot's base.

In this paper we try to reach a compromise between these two approaches. We 
devise a strategy to evaluate if a foothold yields a dynamic transition 
feasibility for the duration of the stance phase of a leg, 
coupled with a vision-based foothold selection that reduces the chance of 
reaching kinematic limits and collisions. We based the 
evaluation on the \gls{ccroc}, presented in \cite{CCROC}. The method looks for 
the existence of feasible trajectories for the \gls{com} (parameterized as B\'ezier curves) 
according to the \gls{rbdm}. Additionally, we implement the method presented in \cite{CCROC} to account for the body angular dynamics in a convex (based on the insight provided in \cite{CCROC}) and a nonlinear fashion. We propose and compare the two different formulations to drop the constant angular momentum assumption: one that keeps the convexity of the optimization problem, and a 
second one that includes the angular dynamics in a nonlinear fashion. We also 
integrate the \gls{vfa} presented in \cite{Villarreal} to discard 
footholds that may lead to collisions or kinematically unreachable 
locations.

The main contributions are:
\begin{enumerate}
	\item A dynamic transition feasibility foothold evaluation that considers linear and angular dynamics of the \gls{rbdm}. We implemented in simulation and experiments two formulations that consider \gls{com} motion and base orientation; a convex one (based on \cite{CCROC}) that includes angular dynamics without breaking convexity and a novel nonlinear approach. 
	
	\item A comparison between a convex and a nonlinear formulation for different scenarios in terms of quality of the generated trajectories.
\end{enumerate}

The paper is organized as follows: \mbox{Section \ref{sec:related_work}} 
summarizes the relevant work; \mbox{Section 
\ref{sec:CTF_formulation}} details the formulation of the dynamic transition 
feasibility and highlights the differences with respect to \cite{CCROC}; in 
\mbox{Section \ref{sec:CTF_solution}} the two formulations of the optimization 
problem are presented; simulation and experimental results 
are presented in \mbox{Section \ref{sec:results}}, and in \mbox{Section 
\ref{sec:conclusion}} we address conclusions and future work.
\section{Related work}\label{sec:related_work}
We focus on legged locomotion strategies that consider variations on the terrain, making a distinction between coupled and decoupled approaches. Furthermore, we place the work here presented as an intermediate solution between both of these categories. 

Coupled approaches control the motion of 
the robot by formulating a single optimal control problem. This means that the 
optimization is posed to find a reference (position and orientation) for the 
body, contact locations (footholds), and inputs (\glspl{grf}, torques), for a defined planning horizon. One of the most remarkable examples is the one proposed by Winkler et 
al. \cite{Winkler2018} which makes the problem more tractable by 
modeling the robot by means of the \gls{rbdm}, although the optimization is still a 
\gls{nlp}. In this case, not only reference trajectories for the body and 
\glspl{grf} are optimized, but also gait timings are included as decision 
variables. The richness of the optimized motions is demonstrated in complex 
scenarios. Similar approaches to this were presented previously by solving the 
optimization problem via \gls{slq} \cite{Neunert} and fixing the gait sequence 
to only optimize timing, posing the problem as a switched system 
\cite{farbod17icra}. A different approach is taken by Aceituno et al. 
\cite{aceituno18ral}, where the proposed algorithm computes gait pattern, contact 
sequence, and \gls{com} trajectories as an outcome of a \gls{micp} on several 
convex surfaces. All of the previously mentioned coupled approaches 
showcase the advantages and the versatility of the generated motions by 
optimizing inputs, body references and footholds together. However, all of 
these suffer from large computational times and 
risk getting stuck in local minima.

Decoupled approaches outsource the foothold selection to an
external module. This reliefs computational cost from 
the optimization, since the foothold positions have a nonlinear relationship 
with the \gls{com} position \cite{Orin}. A method that paved the 
way to account for the terrain was proposed by Kalakrishnan et al. 
\cite{Kalakrishnan}, where terrain was discretized considering templates 
corresponding to portions of the map in the vicinity of a nominal landing 
position. A linear regression method was used to approximate the selection of 
an expert user to adapt the landing location of the feet within the template 
and the motion of the base was designed to follow these footholds. Inspired by 
this, methods have resourced to template-based foothold selection using 
learning-based \cite{Villarreal2,Villarreal,esteban20clawar,barasuol15iros,Belter2,Chen} or fast 
optimization \cite{Fankhauser2,Jenelten} strategies. In this way the optimization 
problem becomes lighter. The main caveat when selecting footholds prior to the motion 
is that in general only geometric constraints are considered such as 
collisions, terrain roughness or kinematic limits. This situation might lead to 
postures in which the robot is not able to continue the motion due to dynamic 
infeasibility (e.g., the motion is not achievable due to actuation limits).

Fernbach et al. \cite{CCROC} presented a method to evaluate the transition 
feasibility of a motion with contact switches. The method relies on the 
parameterization of the trajectory of the \gls{com} as a B\'ezier curve, which 
allows to pose the problem as a quadratic program. The method makes use of the \gls{rbdm}. Tsounis et al. \cite{Tsounis} 
used this method to learn dynamic transition feasibility of foothold and 
devised a learning-based locomotion strategy to generate locomotion in 
complex scenarios with variable terrain. In a similar line, the authors of \cite{Lin} trained  
two coupled neural networks to evaluate the feasibility of the contacts and 
generate the motion trajectory based on the \gls{micp} solved in 
\cite{aceituno18ral}.

We extend the work of 
\cite{CCROC} to evaluate transition feasibility considering variations of 
the angular dynamics with both convex and nonlinear formulations, and on the other hand, we use the found trajectory of the 
\gls{com} and base orientation as reference to guide the base to provide a 
consistent motion with the transition feasibility metric.

\section{Formulation of Contact Transition Feasibility with Angular Dynamics} \label{sec:CTF_formulation}
We address the evaluation of dynamic trajectories for the duration of the stance 
phase of a specific leg. In other words, we want to assert the existence of 
feasible trajectories in a receding horizon fashion. We generate these 
trajectories according to the \gls{rbdm} considering as a basis 
the work of \cite{CCROC} and account for 
the angular dynamics in the optimization. It is worth noticing that in \cite{CCROC}, a theoretical implementation of $\dot{\mathbf{L}} \neq 0$ is discussed. We propose in the next sections two ways to address this consideration.

We follow the same argument given in \cite{CCROC}, which is to connect two 
different sets of states in space and time, we include angular quantities to the set of states. Thus, a 
state is defined as ${\mathbf{x}(t) = [\mathbf{c}(t) \; \mathbf{\dot{c}}(t) \; 
\mathbf{\ddot{c}}(t) \; \bm{\Theta}(t) \; \dot{\bm{\Theta}}(t) \; 
\ddot{\bm{\Theta}}(t)]^\intercal \in \mathbb{R}^{6 \times 3}}$, where 
$\mathbf{c}$ is the position of the \gls{com} and $\bm{\Theta}$ is the orientation of 
the base, expressed in terms of roll ($\phi$), pitch ($\gamma$) and yaw 
($\psi$) angles.

A feasible dynamic transition subject to dynamics constraints that connects two sets of states is defined as ${\bm{f}(t): \mathbf{x}(t_0) \rightarrow \mathbf{x}(t_f),\; t_f > t_0}$, 
where $t_0$ and $t_f$ correspond
to the initial and final states respectively. We employ continuously differentiable, parametric curves to describe position and orientation (similarly to \cite{CCROC}) to reduce the number of decision variables.
\subsection{Time horizon description}
We choose to express the considered time horizon in terms of \textit{contact 
 switches} to have a general description, applicable to any type of gait. A 
 contact switch happens whenever any foot makes or breaks a contact with the 
 ground. 
We define a \gls{csh} as the number of non-simultaneous contact switches 
occurring in the considered period $T = t_f - t_0$. 
\begin{figure}[t]
\centering
\begin{tikzpicture}[scale=0.6]
	\node [draw, circle, inner sep=1.5pt, label=left:{$\mathbf{x}_0$}] (CoM_0) at (0, 0) {};
	\node [draw, circle, inner sep=1.5pt] (CoM_1) at (1, 0.3) {};
	\node [draw, circle, inner sep=1.5pt] (CoM_2) at (1.8, 0.7) {};
	\node [draw, circle, inner sep=1.5pt] (CoM_3) at (3, 0.4) {};
	\node [draw, circle, inner sep=1.5pt, label=right:{$\mathbf{x}_f$}] (CoM_4) at (4, 0.7) {};
	
	\draw [thick] (CoM_0)
	.. controls (0.3, -0.3) and (0.7, -0.1) ..
	(CoM_1) node[pos=0.5, label=above:{1}]{};
	\draw [thick] (CoM_1)
	.. controls (1.3, 0.8) and (1.5, 1) ..
	(CoM_2) node[pos=0.5, label=above:{2}]{};
	\draw [thick, dotted] (CoM_2)
	.. controls (2.2, 0.2) and (2.4, 0.7) ..
	(CoM_3) node[pos=0.7, label=above:{$\ldots$}]{};
	\draw [thick] (CoM_3)
	.. controls (3.3, 0.2) and (3.8, 0.3) ..
	(CoM_4) node[pos=0.5, label=above:{$i$}]{};
\end{tikzpicture}
\caption{Example of sub-horizon partitioning. The CoM trajectory connecting the initial ($\mathbf{x}_0$) and the final ($\mathbf{x}_f$) states is partitioned into multiple sub-horizons. Each of them is labeled according to an increasing index $i$ and is subject to a different set of constraints.}
\label{fig:horizon_partitioning}
\end{figure}
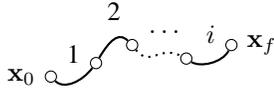
Once a desired evaluation time horizon is chosen, we compute the \gls{csh} 
assuming a periodic gait. A similar approach has been adopted in 
\cite{CCROC}, where the trajectory is split at contact switches, to consider 
transitions between different phases.

For the sake of completeness, we provide a brief description of the method 
presented in \cite{CCROC} pointing out the main differences with respect to 
what is presented here. In \cite{CCROC}, B\'ezier curves are adopted 
to describe the \gls{com} trajectory. They are curves
parametrized by \textit{control points}. Thanks to the properties of 
B\'ezier curves, one can ensure that the generated continuous trajectories 
remain within the constraints. 
In \cite{CCROC}, it is mentioned that the number of contact switches can be 
increased arbitrarily. However, in our experience, we faced feasibility 
problems since the method generates a single B\'ezier curve for the 
entire horizon (with multiple contact switches) with only one degree of freedom (a single free moving control 
point). The solution space of feasible trajectories is thus reduced 
as the number of contact switches increases, since the number of constraints 
increases. To prevent such reduction, we adopt a \gls{csh} partitioning method (\textit{sub-horizon}, Fig. \ref{fig:horizon_partitioning}). We 
choose to limit each sub-horizon to two contact switches to prevent the reduction 
of the solution space. Such limit is specifically chosen to connect two sub-sequent stance 
phases, i.e., lift-off and touchdown of the same leg. Then, to evaluate an arbitrarily large 
\gls{csh}, we concatenate multiple sub-horizons by means of continuity constraints. This leads to a set of parametric 
curves, one curve for each sub-horizon, chained together at \textit{way-points} (connection points between sub-curves) to evaluate a longer \gls{csh}. 

We adopt the \gls{rbdm} to assert dynamic transition feasibility:
\begin{equation}
	\underbrace{\begin{bmatrix}
		m(\ddot{\mathbf{c}} - \mathbf{g}) \\
		m \mathbf{c} \times (\ddot{\mathbf{c}} - \mathbf{g}) + \dot{\mathbf{L}}
	\end{bmatrix}}_{\mathbf{w}}=
	\underbrace{\begin{bmatrix}
		\mathbf{I}_3 & \ldots & \mathbf{I}_3\\
		[\mathbf{p}_1]_\times & \ldots & [\mathbf{p}_j]_\times
	\end{bmatrix}}_{\mathbf{A}}
	\mathbf{f}
\end{equation}
where $\mathbf{g}$ is the gravity vector, $\mathbf{I}_3 \in \mathbb{R}^{3 
\times 3}$ is the identity matrix, $\mathbf{f} =[\mathbf{f}_1 \text{  
}\ldots\text{ }\mathbf{f}_j]^\intercal$, $\mathbf{f}_j \in \mathbb{R}^3$ is the 
ground reaction force (GRF) associated to the j$^{th}$ at point 
$\mathbf{p}_j \in \mathbb{R}^3$ expressed in the world frame and $m$ is the 
robot's mass. 
We express 
$\dot{\mathbf{L}}$ as a function of the angular quantities (orientation, rate 
and acceleration) by analytically differentiating $\mathbf{L} = 
\mathbf{I}_{\mathcal{W}}\bm{\upomega}$, where $\bm{\upomega}$ is angular 
velocity of the rigid body expressed in the CoM frame. One can express 
$\bm{\upomega}$ in terms of $\dot{\bm{\Theta}}$ making use of the following 
mapping:
\begin{align}
\bm{\upomega} = \mathbf{T}(\gamma, \psi) \dot{\mathbf{\Theta}}
\end{align}
where the matrix $\mathbf{T}$ is the matrix that maps angular velocity in 
the world frame to rotation rates.
The derivative of $\mathbf{L}$ is given by:
%
\begin{align} \label{eqn:L_dot}
	\dot{\mathbf{L}} =& \underbrace{\mathbf{T} \bm{\dot \Theta} \times 
	\mathbf{I}_{\mathcal{W}} 
	\mathbf{T}\dot{\bm{\Theta}}}_{\dot{\mathbf{I}}_{\mathcal{W}}\bm{\upomega}}  
	+ \underbrace{\mathbf{I}_{\mathcal{W}}\cdot(\dot{\mathbf{T}} 
	\dot{\bm{\Theta}} + 
	\mathbf{T}\ddot{\bm{\Theta}})}_{\mathbf{I}_{\mathcal{W}}\dot{\bm{\upomega}}}
\end{align}
Note that \eqref{eqn:L_dot} is a highly nonlinear expression that depends on $\bm{\Theta}$ and its derivatives.


In the following section we describe the two proposed formulations (convex and 
nonlinear) to solve the dynamic transition feasibility problem while accounting for 
the rate of change of angular momentum $\dot{\mathbf{L}}$. The first formulation aims at preserving the 
convexity of the problem, thus making it computationally efficient and not 
prone to local minima, at the cost of a more limited solution space. The second 
formulation offers a wider solution space with more computational cost.
\section{Solution of Contact Transition Feasibility with Angular Dynamics} \label{sec:CTF_solution}
\begin{figure}[t]
\centering
\begin{tikzpicture}
	\node [draw, circle, inner sep=1.5pt, label=left:{$\mathbf{c}_0$}] (CoM_0) at (0, 0) {};
	\node [draw, circle, inner sep=1.5pt] (CoM_1) at (1, 0.3) {};
	\node [draw, circle, inner sep=1.5pt] (CoM_2) at (1.8, 0.7) {};
	\node [draw, circle, inner sep=1.5pt, , label=right:{$\mathbf{c}_f$}] (CoM_3) at (2.8, -0.1) {};
	
	\draw [thick] (CoM_0)
	.. controls (0.3, -0.3) and (0.7, -0.1) ..
	(CoM_1);
	\draw [thick] (CoM_1)
	.. controls (1.3, 0.8) and (1.7, 1) ..
	(CoM_2);
	\draw [thick] (CoM_2)
	.. controls (2.2, 0.2) and (2.5, 0) ..
	(CoM_3);
\end{tikzpicture}
\begin{tikzpicture}
	\node [draw, circle, inner sep=1.5pt, label=left:{$\mathbf{c}_0$}] (CoM_0) at (0, 0) {};
	\node [draw=blue, dashed, circle, inner sep=1.5pt] (CoM_1_old) at (1, 0.3) {};
	\node [draw=blue, dashed, circle, inner sep=1.5pt] (CoM_2_old) at (1.8, 0.7) {};
	\node [draw=blue, dashed, circle, inner sep=1.5pt] (CoM_3_old) at (2.8, -0.1) {};

	\node [draw, circle, inner sep=1.5pt] (CoM_1) at (0.8, 0.4) {};
	\node [draw, circle, inner sep=1.5pt] (CoM_2) at (2, 0.6) {};
	\node [draw, circle, inner sep=1.5pt, label=right:{$\mathbf{c}_f$}] (CoM_3) at (2.6, 0.3) {};
	
	\draw [thick] (CoM_0)
	.. controls (0.3, -0.1) and (0.5, -0.1) ..
	(CoM_1);
	\draw [thick] (CoM_1)
	.. controls (1.1, 1) and (1.7, 1) ..
	(CoM_2);
	\draw [thick] (CoM_2)
	.. controls (2.2, 0.4) and (2.2, 0.3) ..
	(CoM_3);
\end{tikzpicture}
\caption{Example of \textit{convex} (left) and \textit{nonlinear} (right) CoM linear trajectories. The CSH is divided in different parts with their own set of constraints. The initial ($\mathbf{c}_0$), final ($\mathbf{c}_f$) and intermediate positions have to be defined a priori in the convex case, while in the nonlinear one they are optimized. The solid black circles are the computed states, while the blue dashed circles are the desired states before the corrective action of the nonlinear optimization.}
\label{fig:convex_img}
\end{figure}
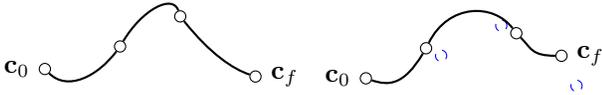

In this section, we describe two formulations to solve the transition 
feasibility problem, namely the \textit{convex} and the \textit{nonlinear} 
approaches. In the convex approach, we overcome the nonconvexity of the 
$\dot{\mathbf{L}}$ analytical expression by decoupling the linear and the 
angular parts and including $\dot{\mathbf{L}}$ as an optimization variable. 
This formulation yields a more limited solution space with respect to the nonlinear one. 
This is because the intermediate points in the CSH are fixed, whereas in the nonlinear case we allow variations of the intermediate and final desired states, as shown in Fig. \ref{fig:convex_img}.
To do so \eqref{eqn:L_dot} as a constraint in function of angular quantities ($\bm{\Theta}$, 
$\dot{\bm{\Theta}}$, $\ddot{\bm{\Theta}}$) which are now included as decision variables, leading to smoother and more versatile motions.

To produce consistent solutions, we include boundary constraints in our formulation.
We define a pair of 
initial and final states associated to an i$^{th}$ sub-horizon $\mathbf{x}_{0, 
\; i}$, $\mathbf{x}_{f, \; i}$ (Fig. \ref{fig:horizon_partitioning}). $\mathbf{x}_{0, 
\; i} = \mathbf{x}_{f, \; i-1}$ is the definition of a continuous function. We can write this 
relationship for every sub-horizon as:

\begin{align}
\left[\begin{matrix}
\mathbf{x}_{0,\; 2}&
\mathbf{x}_{0,\; 3}&
\ldots &
\mathbf{x}_{0,\; i}
\end{matrix}\right]^\intercal
&=
\left[\begin{matrix}
\mathbf{x}_{f, \; 1}&
\mathbf{x}_{f, \; 2}&
\ldots &
\mathbf{x}_{f, \; i - 1}
\end{matrix}\right]^\intercal\nonumber\\
\mathbf{X}_0 &= \mathbf{X}_f
\end{align}
This relationship is implicitly valid for all the sub-horizons in the convex 
formulation because of predefined states, but in the 
nonlinear case it has to be explicitly enforced.
\subsection{Convex formulation}
\label{subsect:convex}

For each sub-horizon we parameterize the \gls{com} trajectory with an 
8$^{th}$ order B\'ezier curve to define the trajectory up to its 
third analytical derivative. A generic n$^{th}$ order B\'ezier curve has 
$(n+1)$ control points and each can be associated to a 
different state quantity (position, velocity and acceleration). We then leave a free control point per each sub-curve.
The collected points are used as optimization variables
($\bm{\uprho}$) and the system 
dynamics are described as
\begin{align}
	\underbrace{\begin{bmatrix}
			m(\ddot{\mathbf{c}}(\bm{\uprho}) - \mathbf{g}) \\
			m \mathbf{c}(\bm{\uprho}) \times (\ddot{\mathbf{c}}(\bm{\uprho}) - 
			\mathbf{g}) + \dot{\mathbf{L}}
	\end{bmatrix}}_{\mathbf{w}(\bm{\uprho})} &=
	\underbrace{\begin{bmatrix}
			\mathbf{I}_3 & \ldots & \mathbf{I}_3 \nonumber\\
			[\mathbf{p}_1]_\times & \ldots & [\mathbf{p}_j]_\times
	\end{bmatrix}}_{\mathbf{A}}
	\mathbf{f}\\
	\mathbf{w}(\bm{\uprho}) &= \mathbf{A} \mathbf{f}
\end{align}
where $\bm{\uprho} \in \mathbb{R}^{i \times 3}$ is the collection of free 
control points associated to each sub-horizon (Fig. 
\ref{fig:horizon_partitioning}). Although the term $\mathbf{c}(\bm{\uprho}) \times \ddot{\mathbf{c}}(\bm{\uprho})$ is a nonlinear term, it can be shown that it can be formulated as a linear constraint by employing B\'ezier curves to describe the CoM trajectory $\mathbf{c}$ \cite{CCROC}.

To include the angular momentum rate $\dot{\mathbf{L}}$ preserving convexity, 
we define $\dot{\mathbf{L}}$ as an optimization variable and compute a desired $
\dot{\mathbf{L}}_{ref}$, which is then tracked by including the term 
$\parallel \mathbf{\dot L} - \mathbf{\dot L}_{ref} \parallel^2_2$ in 
the cost function. To compute $\dot{\mathbf{L}}_{ref}$, we define a desired 
angular behaviour of the robot by designing a trajectory for the angular variables 
($\bm{\Theta}$, $\dot{\bm{\Theta}}$, $\ddot{\bm{\Theta}}$) and then 
computing the angular momentum rate with the full expression \eqref{eqn:L_dot}. 
In the convex case, we decide to describe the trajectory of 
the desired $\bm{\Theta}$ as a B\'ezier curve, to keep a finite number of 
parameters. 
We then formulate the optimization problem as:
\begin{subequations}
\begin{alignat}{2}
&\!{\boldsymbol\min_{{\bm{\uprho}, \mathbf{f}, \dot{\mathbf{L}}}}}      
&\displaystyle\sum_{k=0}^{N\cdot i}& \parallel \mathbf{\dot L}_k - \mathbf{\dot 
L}_{ref, \; k} \parallel_2^2 + \parallel \ddot{\mathbf{c}}_k(\bm{\uprho}_i) 
\parallel_2^2 \notag\\
&\text{\textbf{subject to}} &      & \mathbf{w}_k(\bm{\uprho}_i) = \mathbf{A}_k
\mathbf{f}_k\\
&				&		&  0 \; \leq f_{z, \; k} \; \leq f_{max} 
\label{eqn:f_z_constraint_convex}\\
&				&		&  |f_{x, \; k}| \leq \mu f_{z, \; k},  \quad  |f_{y, \; k}| \leq \mu f_{z, \; k} 
\end{alignat}
\end{subequations}
where $f_{max}$ is an upper limit for the $z$ direction of the force that the robot can exert on the 
ground. The tracking cost for the desired angular 
momentum rate is given by $\parallel \mathbf{\dot L} - \mathbf{\dot L}_{ref} 
\parallel_2^2$ and we minimize the accelerations ($\parallel 
\ddot{\mathbf{c}}(\bm{\uprho}) \parallel_2^2$) to incentivize
smoother trajectories. This first method of including $\dot{\mathbf{L}} \neq 0$ differs from \cite{CCROC} since we are not directly including $\dot{\mathbf{L}}$ as a parametric curve, but rather expressing $\dot{\mathbf{L}}$ as a function of a desired angular trajectory. 
\subsection{Nonlinear formulation}
We present an alternative formulation to the one presented in Section 
\ref{subsect:convex}. The main difference is that herein we aim to 
directly optimize the trajectories of the angular quantities along the 
\gls{csh}, instead of using them as parameters to generate and track a desired $\dot{\mathbf{L}}$. 
One of the main advantages of this approach is that this method does not 
require to set extra constraints to maintain physical consistency. In particular we 
introduced slack variables  on the position and acceleration of the 
intermediate and final states, keeping fixed the velocity to reach the 
commanded velocity at each way-point (Fig. \ref{fig:convex_img}). This makes the problem 
nonlinear due to the dependency of $\dot{\mathbf{L}}$ with respect to the 
angular position, velocity and acceleration. We then define a set of 
\textit{enhanced states} as $\overline{\mathbf{x}} = \mathbf{x}_d + \mathbf{x}_v$, composed by a desired $\mathbf{x}_d = \left[\begin{matrix}
		\mathbf{c}_d&
		\dot{\mathbf{c}}_d&
		\ddot{\mathbf{c}}_d&
		\bm{\Theta}_d&
		\dot{\bm{\Theta}}_d&
		\ddot{\bm{\Theta}}_d
	\end{matrix}\right]^\intercal$ and a variable ${\mathbf{x}_v = \left[\begin{matrix}
		\Delta \mathbf{c}&
		0&
		\Delta \ddot{\mathbf{c}}&
		0&
		0&
		0
	\end{matrix}\right]^\intercal}$ part,
\begin{figure*}[tb]
\centering
\def\svgwidth{0.8\textwidth}
    \adjustbox{trim=0.0cm 0.0cm 0.0cm 
    	0.6cm}{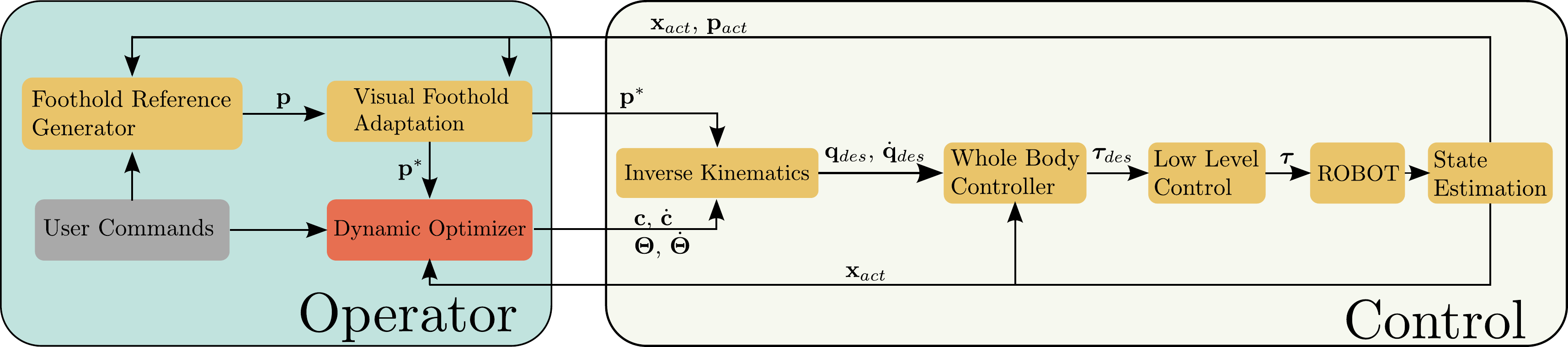}
\caption{Locomotion control pipeline used to implement the proposed foothold evaluation strategies in both experiments and simulations.} 
\label{fig:control_scheme}
\end{figure*}
where ${\Delta \mathbf{c} \in \mathbb{R}^3}$ and $\Delta \ddot{\mathbf{c}}_v \in 
\mathbb{R}^3$ are decision variables.

We consider velocity and position/orientation variables as commanded quantities. Although they are not enhanced, they are 
still allowed to vary in between user-defined states in order to reach the 
designated value at each way-point while fulfilling the imposed constraints. 
The optimization problem is then built as follows:
\begin{subequations}
	\begin{alignat}{2}
		&\!{\boldsymbol\min_{{\bm{\uprho}, \mathbf{f}, \Delta \mathbf{c},\Delta 
		\ddot{\mathbf{c}}, \bm{\Theta}, \dot{\bm{\Theta}}, 
		\ddot{\bm{\Theta}}}}}      &\quad&\displaystyle\sum_{k=0}^{N\cdot i} \parallel \dot{\mathbf{L}_k} 
		\parallel_2^2 + \parallel \ddot{\mathbf{c}}_k(\bm{\uprho}_i, \; \Delta 
		\ddot{\mathbf{c}}_i) \parallel_2^2 
		\label{eqn:min_prob}\\
		&\text{\textbf{subject to}} &      & \mathbf{w}_k(\bm{\uprho}_i, \Delta 
		\mathbf{c}_i, \Delta \ddot{\mathbf{c}}_i) = \mathbf{A}_k \mathbf{f}_k\\
		&				&		& {\dot{\mathbf{L}}_k
		= \dot{\mathbf{L}}_f(\bm{\Theta}_k, \dot{\bm{\Theta}}_k, 
		\ddot{\bm{\Theta}}_k)} \label{eqn:L_dot_constraint}\\
		&				&		& \mathbf{X}_0 = \mathbf{X}_f\\
&				&		&  0 \; \leq f_{z, \; k} \; \leq f_{max} 
\label{eqn:f_z_constraint_convex}\\
&				&		&  |f_{x, \; k}| \leq \mu f_{z, \; k},  \quad  |f_{y, \; k}| \leq \mu f_{z, \; k}
	\end{alignat}
\end{subequations}
where the cost function is similar to the one adopted in the convex 
formulation. The main difference consists in $\parallel \dot{\mathbf{L}_k}
\parallel_2^2$, which is a cost term that aims at reducing the angular 
variation rather than tracking a reference behavior, helping to incentivize 
less aggressive motions for the angular quantities. We then 
consider \eqref{eqn:L_dot} as a constraint dependent in the angular quantities 
defined as $\dot{\mathbf{L}}_f(\bm{\Theta}, \dot{\bm{\Theta}}, 
\ddot{\bm{\Theta}})$ and account for these quantities as decision variables.
This formulation is computationally more expensive than the convex formulation
and it is prone to local minima, but the output trajectories are qualitatively better compared to the convex
formulation solutions. 

\section{Results}

\begin{figure}[t]
\centering
\includegraphics[width=0.2\textwidth, trim=80 0 90 40, clip]{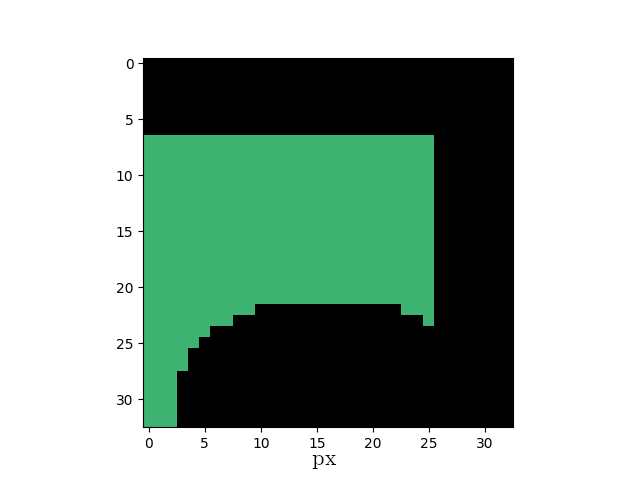}%
\includegraphics[width=0.255\textwidth, trim=55 0 35 40, clip]{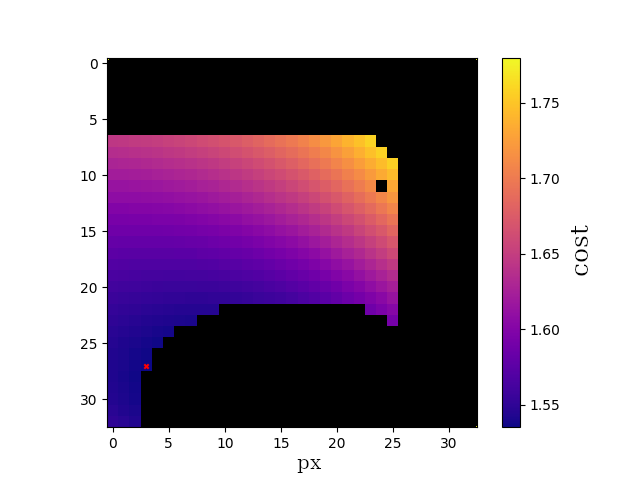}%
\caption{Comparison of \gls{vfa} (terrain roughness, kinematic reachability, collisions) versus dynamic foothold evaluation for the RF foot during stair climbing. In the case of the \gls{vfa} evaluation, green pixels indicate safe landing locations and black unsafe. For the dynamic evaluation black pixels are dynamically infeasible footholds and we show the cost from the solution of \eqref{eqn:min_prob} indicated by the colorbar.}
\label{fig:feas_patch}
\end{figure}

\label{sec:results}
In this section we evaluate the proposed evaluation criterion and the two proposed formulations. We show an example of an evaluation of a series of contact locations for foothold selection and verify the feasibility of the generated trajectories in a stair climbing scenario. Additional results on flat terrain are reported in the accompanying video.

\subsection{Implementation details}
\label{sec:implementation_details}
To verify the feasiblity of the generated trajectories provided by the foothold evaluation, we adopt the control scheme shown in Fig. \ref{fig:control_scheme}. The 
\gls{frg} is in charge of generating the foothold positions with respect to the 
actual \gls{com}['s] states and foothold positions ($\mathbf{x}_{act}$, 
$\mathbf{p}_{act}$), according to the user commanded velocity. 
The reference foothold positions are evaluated with the \gls{vfa} 
\cite{Villarreal}, 
which discards unsuitable footholds according to geometric constraints and 
sends the 
adapted foothold positions $\mathbf{p}^*\in \mathbb{R}^{nc \times 3}$ to both the Dynamic Optimizer and IK blocks, for all 
the contact points $nc$. The 
Dynamic Optimizer block encapsulates the methods presented in this paper. 
Given the set 
of the actual states, the adapted footholds and the user command, it generates 
the state references that will be provided to both the \gls{ik} 
block and the whole-body controller (WBC) \cite{focchi2016}. The \gls{ik} computes 
the joint positions and velocities, providing the \gls{wbc} with a set of joint  
references for the swing leg trajectories. The \gls{wbc} computes the required \glspl{grf} 
to track the 
reference 
quantities, transforming them into joint torques. They are then commanded to 
the 
low-level control, which is in charge of performing the joint torque control.

The convex formulation makes use of the CVXPY modeling 
language \cite{diamond2016cvxpy, agrawal2018rewriting}, with ECOS \cite{ECOS} 
as solver. The nonlinear approach relies on CasADi 
\cite{Andersson2019}, acting as an interface with 
IPOPT \cite{IPOPT}. We evaluated our methods in simulation and experiments. The 
simulations were performed on a i7-8700 CPU 
using Gazebo \cite{gazebo}. In the experimental setup, we execute the 
optimization 
on an off-board computer. The generated reference is sent to the onboard 
Control computer 
via ROS messages \cite{ros}. In the case of the simulations, we 
assume 
no computational time constraints to visualize the behavior of the robot while 
executing the generated trajectories, whereas in the experiment we perform a 
new optimization at each step due to the time constraints of the approach.
\subsection{Foothold Dynamic Feasibility Evaluation}

\begin{figure}[t]
\centering
\def\svgwidth{0.9\columnwidth}
    \adjustbox{trim=0.2cm 0.2cm 0.3cm 
    0.0cm}{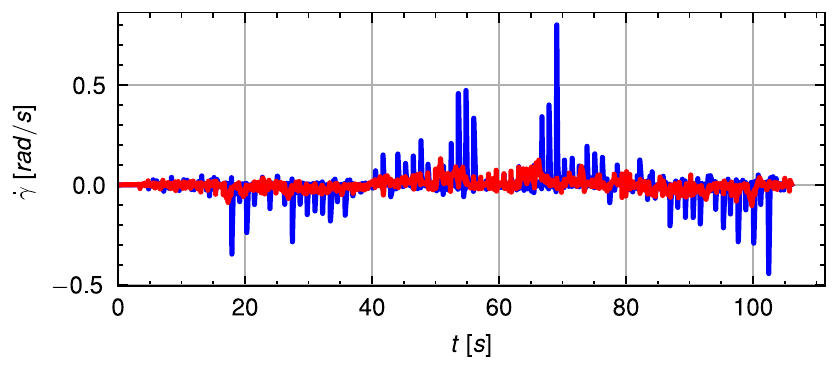}
\captionsetup{belowskip=0.0pt}
\caption{Comparison between angular rates (pitch rate) in simulated 
stairs scenario. The solid blue line is the pitch rate
traversing the scenario considering $\dot{\mathbf{L}} = 0$, while the solid red 
line is the angular velocity achieved using our nonlinear approach, where 
$\dot{\mathbf{L}} \neq 0$.}
\label{fig:ang_compare}
\end{figure}

We evaluate the dynamic transition feasiblity on a stair climbing scenario. We compute a set of nominal future foothold locations assuming a periodic gait during one gait cycle. We then evaluate the neighboring area of the nominal footholds to avoid reaching kinematic limits, collisions or unsafe landing locations using the \gls{vfa} \cite{Villarreal}. Subsequently, we check if there exists a trajectory that solves the problem formulated in \eqref{eqn:min_prob} for every candidate foothold. If the solution exists, the foothold is deemed dynamically feasible. Additionally we compute a cost map to visualize the "quality" of the solution provided by each candidate foothold. Fig. \ref{fig:feas_patch} shows an example of evaluated candidate footholds for the RF leg during a crawl while climbing stairs. Each pixel in the figure represents a candidate landing location. It can be seen that in the dynamic evaluation (Fig.~\ref{fig:feas_patch} on the right) the footholds located on the top right corner of the area yield a higher cost, and some footholds deemed feasible using the \gls{vfa} (Fig.~\ref{fig:feas_patch} on the left) are discarded since no feasible trajectories to solve \eqref{eqn:min_prob} was found. 

\subsection{Simulations}
\noindent \textbf{Effect of time varying angular momentum.}

We highlight the importance of 
considering $\dot{\mathbf{L}} \neq 0$. Fig. \ref{fig:ang_compare} shows a 
comparison of the pitch velocity for the case where $\dot{\mathbf{L}} = 0$ and 
$\dot{\mathbf{L}} \ne 0$. Both of these while climbing stairs 
using the nonlinear formulation. It can be noted that in 
the case of $\dot{\mathbf{L}} = 0$ the peaks in velocity are considerably 
larger. The same simulation applying the convex version of the approach yielded 
similar results, which are omitted for the sake of brevity. These results 
highlight the importance of the angular dynamics when dealing with rough 
terrain.
\begin{figure}[!b]
\centering
	\def\svgwidth{\columnwidth}
	\adjustbox{trim=0.2cm 0.2cm 0.25cm 
	0.35cm}{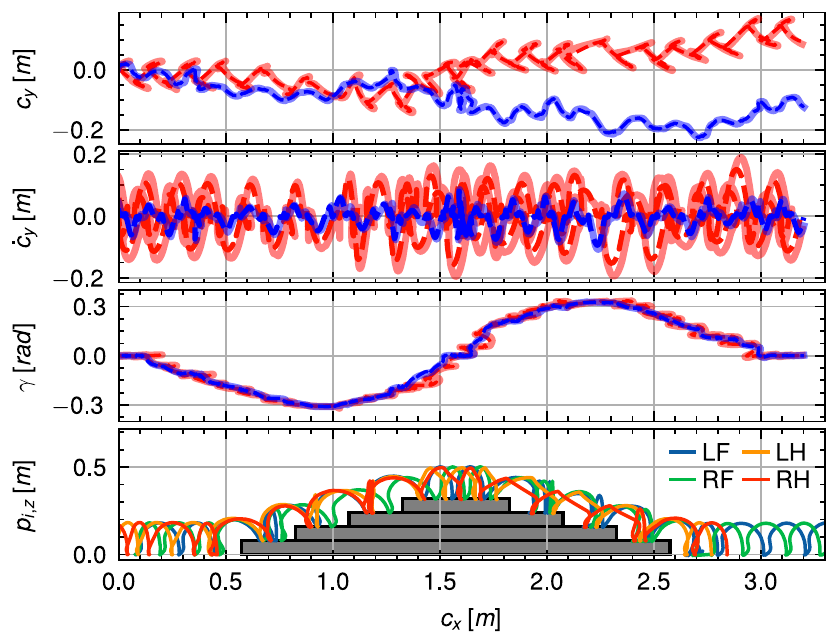}
\caption{Simulation on stairs.	 
The dashed and solid blue lines are the nonlinear reference and tracking, 
respectively. The dashed and solid red lines are the convex
reference and tracking, respectively. Pitch reference and tracking (third row) follow the same color code as linear quantities. In the last 
row an example of feet trajectories is shown, where LF:Left Front, RF:Right 
Front, LH:Left Hind, and RH:Right Hind.}
\label{fig:stairs_perf}
\end{figure}

\noindent \textbf{Comparison between convex and nonlinear formulation.} 

Fig. \ref 
{fig:stairs_perf} shows a comparison of the executed trajectories and their 
reference for both the convex and nonlinear formulation in a stair climbing 
scenario. Although both approaches are able to go up and down the stairs, 
looking closely at the generated trajectory in the $xy$ plane, one can see 
that the nonlinear trajectory yields less aggressive changes of direction 
compared to its convex counterpart. Looking at the velocity in the 
$y$ direction it can be seen that the amplitude of the variation of the 
reference and the executed velocity is always larger in the case of the convex 
formulation with respect to the nonlinear one. We also provide a 
comparison between the two approaches on flat terrain in the accompanying video 
to highlight that even on a simple scenario the nonlinear formulation provides 
smoother trajectories with less peaks in velocity. Regarding computation times, 
the convex took 2.90s for the stair scenario and 2.96s for the flat scenario in 
average over multiple trials to solve the optimization problem, whereas the 
nonlinear 
formulation takes 12.78s for the stair scenario and 8.0s for the flat scenario.
\subsection{Experiments}
\label{sec:experiments}
\begin{figure}[t]
\centering
\def\svgwidth{\columnwidth}
    \adjustbox{trim=0.2cm 0.2cm 0.3cm 0.0cm}{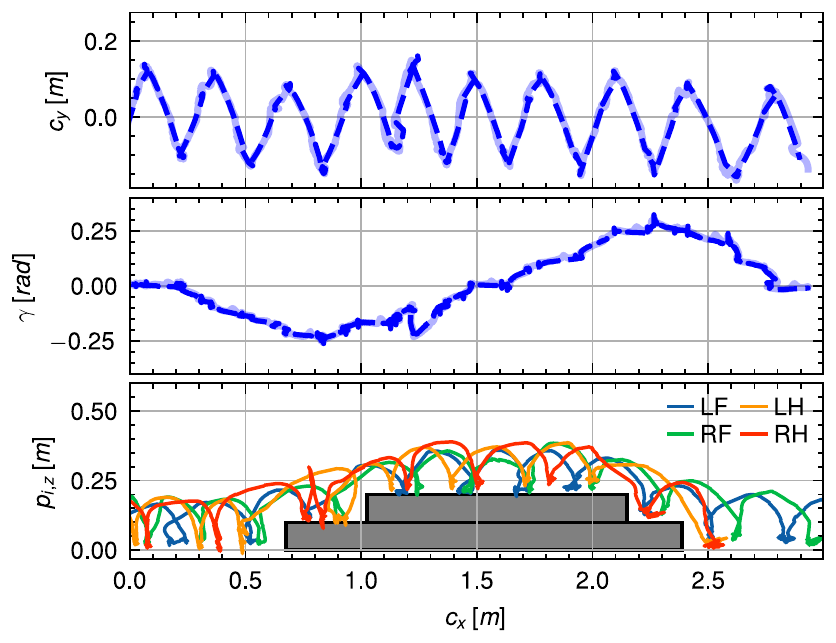}
\caption{Stairs experiment with nonlinear approach. For the top two plots, 
dashed lines are reference and solid lines are tracking signal. The top plot 
shows the CoM trajectory tracking on the $xy$ plane and the middle plot 
shows the pitch tracking. On the last row we 
show the feet trajectories, where LF:Left Front, RF:Right Front, LH:Left Hind, 
and RH:Right Hind.}
\label{fig:experiments}
\end{figure}

In this section we evaluate the hardware feasibility of the optimized trajectories 
on the quadruped robot HyQ. We decided to test the trajectory generated by the
nonlinear formulation since it is the one that proved better in terms of generated trajectories. The scenario tested is 
shown in the snapshots of Fig. 
\ref{fig:hyq_pallet_exp}. 
It consists of climbing and descending two steps of 8cm each. Fig. \ref{fig:experiments} shows the 
tracking of the \gls{com} 
trajectory on the $xy$ plane and the pitch of the robot. As it can be seen, the 
robot is able to cross the scenario with a low tracking error in both position 
and orientation. As in the case of the simulation, the accompanying video shows 
an experiment of the robot executing the optimized trajectories on flat terrain 
and stairs.
\section{Conclusions}\label{sec:conclusion}
We presented a foothold evaluation criterion to assess the 
existence of dynamically feasible trajectories for legged locomotion that 
considers both linear and angular dynamics. We extended the method in 
\cite{CCROC} by formulating the problem allowing variations in the angular 
momentum along the trajectory (i.e., $\dot{\mathbf{L}} \ne 0$) as a function of a desired angular trajectory. 
We presented two different formulations (a convex and a 
nonlinear) both able to generate feasible \gls{com} trajectories. We showed that the 
convex formulation is faster to compute (4 times in average faster than the 
nonlinear) and not subject to local minima, while its nonlinear counterpart is able to generate smoother trajectories.

When dealing with less dynamic gaits, such as 
crawl, the robot can track any desired linear and angular behaviour. But 
when more dynamic gaits are considered (e.g., trot), the optimization problem might not be able to find a solution due to the inability to track linear and angular quantities simultaneously because of \textit{underactuation}. Not being able to 
forecast how close the solution will be to the reference 
$\dot{\mathbf{L}}_{ref}$ in the convex formulation, may lead to unforeseen consequences in terms of 
tracking and trajectory generation. This would require to bound 
$\dot{\mathbf{L}}$ between physically meaningful limits (e.g. the maximum moment that the robot can counteract), which is out of the scope of this paper.%
A key limitation that affects future implementations of the convex formulation is the
need of designing constraints to limit the value $\dot{\mathbf{L}}$ to be physically achievable given actuators limits. In the case of both approaches, the computation times do
not allow them to be used continuously to assess footholds and provide \gls{com} 
reference online. 

As future work we aim to extend the proposed formulations to include more dynamic gaits, such as trot, and to design a learning algorithm that is able to approximate the proposed formulations, reducing computational burden. 

\bibliographystyle{IEEEtran}
\bibliography{Bibliography}
\end{document}

%% file: acronyms.tex
\newacronym{html}{HTML}{hypertext markup languages}
\newacronym{vfa}{VFA}{Vision-based Foothold Adaptation}
\newacronym[plural=CNNs]{cnn}{CNN}{convolutional neural network}
\newacronym{rcf}{RCF}{Reactive Controller Framework}
\newacronym{imu}{IMU}{inertial measurement unit}
\newacronym{mpc}{MPC}{model predictive control}
\newacronym{com}{CoM}{center of mass}
\newacronym[plural=GRFs]{grf}{GRF}{ground reaction force}
\newacronym{zoh}{ZOH}{zero-order hold}
\newacronym[plural=QPs]{qp}{QP}{quadratic program}
\newacronym{rms}{RMS}{root mean square}
\newacronym[plural=CWCs]{cwc}{CWC}{contact wrench cone}
\newacronym{zmp}{ZMP}{Zero-Moment Point}
\newacronym{cop}{CoP}{Center of Pressure}
\newacronym{fwp}{FWP}{Feasible Wrench Polytope}
\newacronym{mit}{MIT}{Massachusetts Institute of Technology}
\newacronym{cmu}{CMU}{Carnegie Mellon University}
\newacronym[plural=CPGs]{cpg}{CPG}{central pattern generator}
\newacronym[plural=ANNs]{ann}{ANN}{artificial neural network}
\newacronym[plural=NNs]{nn}{NN}{neural network}
\newacronym{lip}{LIP}{linear inverted pendulum}
\newacronym{slip}{SLIP}{spring loaded inverted pendulum}
\newacronym[plural=DOFs,longplural={degrees of freedom}]{dof}{DOF}{degree of 
freedom}
\newacronym{wbc}{WBC}{whole-body control}
\newacronym{to}{TO}{trajectory optimization}
\newacronym[plural=NLP]{nlp}{NLP}{nonlinear program}
\newacronym{slq}{SLQ}{sequential linear quadratic}
\newacronym{ddp}{DDP}{differential dynamic programming}
\newacronym{miqcqp}{MIQCQP}{mixed-integer quadratically-constrained quadratic 
program}
\newacronym{micp}{MICP}{mixed-integer convex program}
\newacronym{sqp}{SQP}{sequential quadratic programming}
\newacronym{nmpc}{NMPC}{nonlinear model predictive control}
\newacronym{ilqr}{iLQR}{iterative linear quadratic regulator}
\newacronym{ana}{ANA*}{anytime non-parametric A*}
\newacronym{pd}{PD}{proportional-derivative}
\newacronym{dls}{DLS}{Dynamic Legged Systems lab}
\newacronym{iit}{IIT}{Istituto Italiano di Tecnologia}
\newacronym{haa}{HAA}{hip abduction/adduction}
\newacronym{hfe}{HFE}{hip flexion/extension}
\newacronym{kfe}{KFE}{knee flexion/extension}
\newacronym{lf}{LF}{left-front}
\newacronym{rf}{RF}{right-front}
\newacronym{lh}{LH}{left-hind}
\newacronym{rh}{RH}{right-hind}
\newacronym{isa}{ISA}{Integrated Smart Actuator}
\newacronym[plural=HPUs]{hpu}{HPU}{hydraulic pressure unit}
\newacronym{am}{AM}{additively manufactured}
\newacronym{vo}{VO}{visual odometry}
\newacronym{icp}{ICP}{iterative closest point}
\newacronym{kl}{KL}{Kullback-Leibler}
\newacronym{ccroc}{C-CROC}{Continuous Convex Resolution of Centroidal Dynamic 
Trajectories}
\newacronym{rbdm}{SRBDM}{single rigid body dynamics model}
\newacronym{csh}{CSH}{contact switch horizon}
\newacronym{frg}{FRG}{Foothold Reference Generator}
\newacronym{ik}{IK}{inverse kinematics}

%% file: simulations/block_diag.pdf_tex
\begingroup%
  \makeatletter%
  \providecommand\color[2][]{%
    \errmessage{(Inkscape) Color is used for the text in Inkscape, but the package 'color.sty' is not loaded}%
    \renewcommand\color[2][]{}%
  }%
  \providecommand\transparent[1]{%
    \errmessage{(Inkscape) Transparency is used (non-zero) for the text in Inkscape, but the package 'transparent.sty' is not loaded}%
    \renewcommand\transparent[1]{}%
  }%
  \providecommand\rotatebox[2]{#2}%
  \newcommand*\fsize{\dimexpr\f@size pt\relax}%
  \newcommand*\lineheight[1]{\fontsize{\fsize}{#1\fsize}\selectfont}%
  \ifx\svgwidth\undefined%
    \setlength{\unitlength}{987.83556609bp}%
    \ifx\svgscale\undefined%
      \relax%
    \else%
      \setlength{\unitlength}{\unitlength * \real{\svgscale}}%
    \fi%
  \else%
    \setlength{\unitlength}{\svgwidth}%
  \fi%
  \global\let\svgwidth\undefined%
  \global\let\svgscale\undefined%
  \makeatother%
  \begin{picture}(1,0.2212489)%
    \lineheight{1}%
    \setlength\tabcolsep{0pt}%
    \put(0,0){\includegraphics[width=\unitlength,page=1]{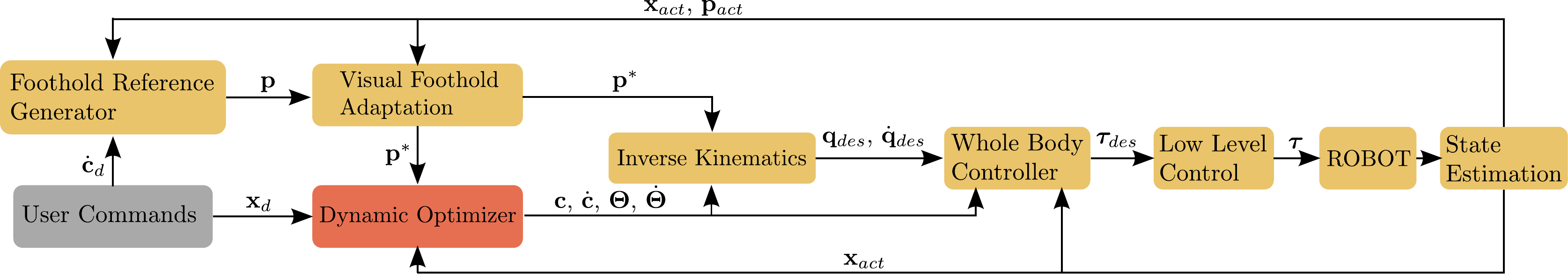}}%
  \end{picture}%
\endgroup%

%% file: simulations/nonlin_pitch_vel.pdf_tex
\begingroup%
  \makeatletter%
  \providecommand\color[2][]{%
    \errmessage{(Inkscape) Color is used for the text in Inkscape, but the package 'color.sty' is not loaded}%
    \renewcommand\color[2][]{}%
  }%
  \providecommand\transparent[1]{%
    \errmessage{(Inkscape) Transparency is used (non-zero) for the text in Inkscape, but the package 'transparent.sty' is not loaded}%
    \renewcommand\transparent[1]{}%
  }%
  \providecommand\rotatebox[2]{#2}%
  \newcommand*\fsize{\dimexpr\f@size pt\relax}%
  \newcommand*\lineheight[1]{\fontsize{\fsize}{#1\fsize}\selectfont}%
  \ifx\svgwidth\undefined%
    \setlength{\unitlength}{241.251134bp}%
    \ifx\svgscale\undefined%
      \relax%
    \else%
      \setlength{\unitlength}{\unitlength * \real{\svgscale}}%
    \fi%
  \else%
    \setlength{\unitlength}{\svgwidth}%
  \fi%
  \global\let\svgwidth\undefined%
  \global\let\svgscale\undefined%
  \makeatother%
  \begin{picture}(1,0.44095063)%
    \lineheight{1}%
    \setlength\tabcolsep{0pt}%
    \put(0,0){\includegraphics[width=\unitlength,page=1]{simulations/nonlin_pitch_vel.pdf}}%
  \end{picture}%
\endgroup%

%% file: simulations/stairs.pdf_tex
\begingroup%
  \makeatletter%
  \providecommand\color[2][]{%
    \errmessage{(Inkscape) Color is used for the text in Inkscape, but the package 'color.sty' is not loaded}%
    \renewcommand\color[2][]{}%
  }%
  \providecommand\transparent[1]{%
    \errmessage{(Inkscape) Transparency is used (non-zero) for the text in Inkscape, but the package 'transparent.sty' is not loaded}%
    \renewcommand\transparent[1]{}%
  }%
  \providecommand\rotatebox[2]{#2}%
  \newcommand*\fsize{\dimexpr\f@size pt\relax}%
  \newcommand*\lineheight[1]{\fontsize{\fsize}{#1\fsize}\selectfont}%
  \ifx\svgwidth\undefined%
    \setlength{\unitlength}{241.20703152bp}%
    \ifx\svgscale\undefined%
      \relax%
    \else%
      \setlength{\unitlength}{\unitlength * \real{\svgscale}}%
    \fi%
  \else%
    \setlength{\unitlength}{\svgwidth}%
  \fi%
  \global\let\svgwidth\undefined%
  \global\let\svgscale\undefined%
  \makeatother%
  \begin{picture}(1,0.7683849)%
    \lineheight{1}%
    \setlength\tabcolsep{0pt}%
    \put(0,0){\includegraphics[width=\unitlength,page=1]{simulations/stairs.pdf}}%
  \end{picture}%
\endgroup%

%% file: simulations/exp.pdf_tex
\begingroup%
  \makeatletter%
  \providecommand\color[2][]{%
    \errmessage{(Inkscape) Color is used for the text in Inkscape, but the package 'color.sty' is not loaded}%
    \renewcommand\color[2][]{}%
  }%
  \providecommand\transparent[1]{%
    \errmessage{(Inkscape) Transparency is used (non-zero) for the text in Inkscape, but the package 'transparent.sty' is not loaded}%
    \renewcommand\transparent[1]{}%
  }%
  \providecommand\rotatebox[2]{#2}%
  \newcommand*\fsize{\dimexpr\f@size pt\relax}%
  \newcommand*\lineheight[1]{\fontsize{\fsize}{#1\fsize}\selectfont}%
  \ifx\svgwidth\undefined%
    \setlength{\unitlength}{240.648903bp}%
    \ifx\svgscale\undefined%
      \relax%
    \else%
      \setlength{\unitlength}{\unitlength * \real{\svgscale}}%
    \fi%
  \else%
    \setlength{\unitlength}{\svgwidth}%
  \fi%
  \global\let\svgwidth\undefined%
  \global\let\svgscale\undefined%
  \makeatother%
  \begin{picture}(1,0.76894894)%
    \lineheight{1}%
    \setlength\tabcolsep{0pt}%
    \put(0,0){\includegraphics[width=\unitlength,page=1]{simulations/exp.pdf}}%
  \end{picture}%
\endgroup%